\documentclass[letterpaper, 10 pt, journal, twoside]{IEEEtran}
\usepackage{times}

\usepackage[numbers]{natbib}
\usepackage{multicol}
\usepackage[bookmarks=true]{hyperref}
\usepackage{graphicx}
\usepackage{physics}
\usepackage{amsmath}
\usepackage{tikz}
\usepackage{mathdots}
\usepackage{yhmath}
\usepackage{cancel}
\usepackage{color}
\usepackage{colortbl}
\usepackage{siunitx}
\usepackage{array}
\usepackage{amssymb}
\usepackage{gensymb}
\usepackage{acronym}
\usepackage{multirow}
\usepackage{tabularx}
\usepackage{extarrows}
\usepackage{booktabs}
\usetikzlibrary{fadings}
\usetikzlibrary{patterns}
\usetikzlibrary{shadows.blur}
\usetikzlibrary{shapes}
\usepackage{subcaption}
\usepackage{cleveref}
\usepackage{algpseudocode}
\usepackage{csquotes}
\usepackage{censor}
\usepackage{soul}
\usepackage{svg}
\let\labelindent\relax
\usepackage{enumitem} 

\graphicspath{{pictures/}}

\acrodef{OG}{Occupancy Grids}
\acrodef{3DSG}{3D Scene Graph}
\acrodef{DoF}{Degree of Freedom}
\acrodef{LLM}{Large Language Model}
\acrodef{BIT}{Batch Informed Trees}
\acrodef{GNN}{Graph Neural Network}
\acrodef{SDF}{Signed Distance Field}
\acrodef{PRM}{Probabilistic Roadmap}
\acrodef{AI}{Artificial Intelligence}
\acrodef{C-space}{Configuration space}
\acrodef{IVS}{Indoor Visibility Structure}
\acrodef{BIM}{Building Information Modeling}
\acrodef{OMPL}{Open Motion Planning Library}
\acrodef{RRT}{Rapidly-Exploring Random Tree}
\acrodef{LiDAR}{Light Detection And Ranging}
\acrodef{ESDF}{Euclidean Signed Distance Field}
\acrodef{SLAM}{Simultaneous Localization and Mapping}

\newcommand{\spath}{S-Path }

\newcommand{\ie}{\textit{i.e., }}
\newcommand{\etal}{\textit{et al. }}

\definecolor{red}{HTML}{fd8f8f}
\definecolor{greend}{HTML}{57e377}
\definecolor{greenl}{HTML}{b8fb8a}
\definecolor{yellow}{HTML}{fefdb4}
\definecolor{orange}{HTML}{ffd5ab}

\begin{document}
% ======================================================================================================

% paper title
\title{Situationally-aware Path Planning \\ Exploiting 3D Scene Graphs \vspace{-1pt}}
% \blackout{
\author{Saad Ejaz, Marco Giberna, Muhammad Shaheer, Jose Andres Millan-Romera,  Ali Tourani, \\Paul Kremer, Holger Voos, and Jose Luis Sanchez-Lopez
    % \thanks{Manuscript received: August 8, 2025; Revised: November 27, 2025; Accepted: January 5, 2026.}
    % \thanks{This paper was recommended for publication by Editor Olivier Stasse upon evaluation of the Associate Editor and Reviewers comments.}
    \thanks{Authors are with the Automation and Robotics Research Group, Interdisciplinary Centre for Security, Reliability and Trust, University of Luxembourg. Holger Voos is also associated with the Faculty of Science, Technology and Medicine, University of Luxembourg, Luxembourg.
    \tt{\small{\{saad.ejaz, marco.giberna, muhammad.shaheer, jose.millan, ali.tourani, p.kremer, holger.voos, joseluis.sanchezlopez\}}@uni.lu}}% 
    \thanks{This work was partially funded by the Fonds National de la Recherche of Luxembourg (FNR) through the DEUS (Ref. C22/IS/17387634/DEUS), INVISIMARK (Ref. DEFENCE22/IS/17800397/INVISIMARK), and RoboSAUR (Ref. 17097684/RoboSAUR) projects; by the Institute of Advanced Studies (IAS) at the University of Luxembourg through the “Audacity” grant (project TRANSCEND - 2021); and by the European Commission euROBIN project (101070596) through the subgrant SATORI (8\_euROBIN\_1OC).
    }
    \thanks{For the purpose of Open Access, the author has applied a CC-BY-4.0 public copyright license to any Author Accepted Manuscript version arising from this submission.}
}
% \markboth{IEEE Robotics and Automation Letters. Preprint Version. Accepted January, 2026}{Ejaz \MakeLowercase{\textit{et al.}}: \textit{Situationally-aware Path Planning Exploiting 3D Scene Graphs}}
\maketitle

\begin{abstract}
3D Scene Graphs integrate both metric and semantic information, yet their structure remains underexploited for improving path planning efficiency and interpretability. 
In this work, we present S-Path, a Situationally-aware Path planner that leverages the metric-semantic structure of indoor 3D Scene Graphs to significantly enhance planning efficiency.
\spath\ follows a two-stage process: it first performs a search over a semantic graph derived from the scene graph to yield a human-understandable high-level path.
This also identifies relevant regions for planning, which later allows the decomposition of the problem into smaller, independent subproblems that can be solved in parallel.
We also introduce a replanning mechanism that, in the event of an infeasible path, reuses information from previously solved subproblems to update semantic heuristics and prioritize re-use to further improve the efficiency of future planning attempts.
Extensive experiments on both real-world and simulated environments show that S-Path achieves average reductions of 6× in planning time while maintaining comparable path optimality to classical sampling-based planners, and surpassing them in complex scenarios,
making it an efficient and interpretable path planner for environments represented by indoor 3D scene graphs.
% Code available at: \hyperlink{https://anonymous.4open.science/r/spath}{https://anonymous.4open.science/r/spath}
Code available at: \hyperlink{https://github.com/snt-arg/spath\_ros}{https://github.com/snt-arg/spath\_ros}

\begin{IEEEkeywords}
Motion and Path Planning, 3D Scene Graphs
\end{IEEEkeywords}
\end{abstract}

\IEEEpeerreviewmaketitle

\vspace{-5pt}
\section{Introduction}
% Intro  
\IEEEPARstart{T}{raditional} path planners for indoor robot navigation rely on dense sampling of the configuration space (C-space) to find feasible paths, but their computational cost grows rapidly with its size and complexity, limiting efficiency and usability.
To mitigate this, some planners  \cite{Gammell2015} incorporate heuristics to sample \textit{intelligently}, but these heuristics often perform poorly in indoor environments characterized by narrow passageways and limited lines of sight.  
Moreover, such planners typically lack semantic understanding, making integration with human-intuitive reasoning and commands challenging.  
To address these limitations, several approaches have incorporated heuristics based on semantic information, such as objects \cite{dharmadhikari2023semantics} and terrain types \cite{zhao2019semantic,ryll2020semantic} into the planning process to improve both efficiency and interpretability.  
However, these methods rely on non-hierarchical, low-level representations, which suffer from scalability and ambiguity issues when applied to large and structurally complex environments.

3D scene graphs \cite{Armeni2019} are emerging as efficient data structures that encode spatial relationships between semantic entities in a hierarchical manner, thereby enabling effective organization of complex spaces \cite{hydra,sgraphsp}.  
\begin{figure}
    \centering
    \includegraphics[width=1.0\columnwidth]{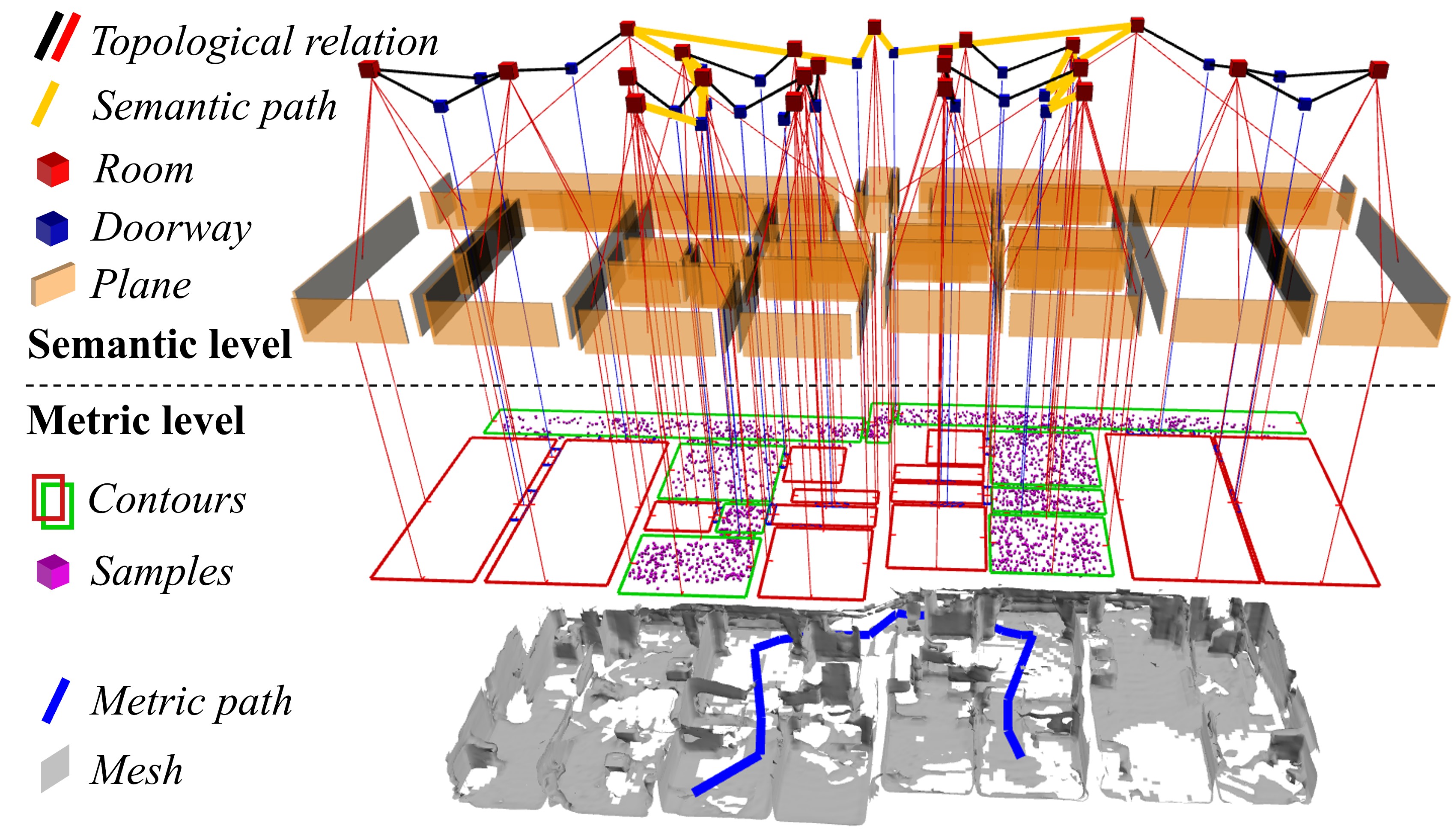}
    \caption{\spath augments sampling-based planners with high-level metric-semantic scene understanding to restrict sampling to only the relevant regions (green contours), thereby reducing planning time in indoor environments. 
    }
    \label{fig:snav-layers}
    \vspace{-5pt}
\end{figure}
Recent research has explored the use of 3D scene graphs for task and motion planning \cite{agia2022taskography, rana2023sayplan, ray2024task}; 
however, they focus solely on plan feasibility, operate predominantly within the task planning domain, and fail to leverage the potential of high-level semantics (e.g., \textit{rooms}, \textit{doorways}, \textit{floors}) to improve path planning efficiency.

\color{ black}
One preliminary work, S-Nav~\cite{kremer2023s}, has explored the idea of using semantic information from scene graphs to decompose a global planning problem into smaller subproblems to improve planning speed. 
However, it is limited to simple, well-defined synthetic environments and lacks any mechanism for efficient replanning to handle evolving situations that commonly arise in real-world scenarios. 
Following this, we introduce S-Path, a \textit{Situationally-aware Path planner} that leverages the structure of indoor scene graphs to achieve substantial gains in planning efficiency. Our planner is designed for real-world, large-scale environments and produces human-interpretable semantic paths while remaining robust through an efficient replanning mechanism.
\color{black}
% Building upon \cite{kremer2023s}, 
S-Path decomposes a global planning problem into smaller, independent subproblems by performing a search over the high-level semantic layers of the scene graph. These subproblems are then solved in parallel at the geometric level using sampling-based planners, which operate significantly faster within the restricted C-spaces yielded by the decomposition.
Moreover, if the planned path is or becomes invalid, \spath can swiftly replan by focusing only on the affected portions of the problem, while reusing previously solved subproblems to efficiently construct a new path.
The main contribution of this paper is a path planner that:
\begin{itemize}
    \item exploits the hierarchical structure of indoor 3D scene graphs to constrain planning to only the relevant regions of the environment.
    \item decomposes the global planning problem into smaller, independent subproblems, enabling parallel solving, thereby reducing planning time.
    \item includes a resource-efficient replanning strategy that isolates and updates only the affected parts of the plan, thereby reducing replanning latency.
\end{itemize}

Moreover, \spath is built as a ROS-based framework utilizing the \ac{OMPL} framework \cite{Sucan2012} and provides a lightweight interface to exploit the semantics of 3D indoor scene graphs to enhance the efficiency of sampling-based planners.

% ======================================================================================================
\section{Related Works}

\subsection{Semantics-aware Path Planning}
Terrain semantics such as the classification of reliable terrain has been employed in path planning to guide UGVs \cite{zhao2019semantic} and UAVs \cite{ryll2020semantic}.
In other works, semantic maps generated using cameras, semantic segmentation, and object detection, have been utilized.
Works, such as \cite{Chaplot2020}, utilize semantic information about various objects in the environment to inform path planning for efficient object search.  
Dharmadhikari \etal~\cite{dharmadhikari2023semantics} presented a semantics-aware path planner that explores unknown environments, reconstructs meshes, and inspects semantics of interest.  
Ryll \etal~\cite{ryll2020semantic} demonstrated how semantic information improves the perception–estimation–planning loop for high-speed drone flight in urban settings.
However, these works incorporate semantic information in a non-hierarchical manner, using it solely to inform path planning decisions rather than to improve planning efficiency.

\color{ black}Moreover, some works use hierarchical semantic abstractions for global-local planning.
% Sun \etal~\cite{sun2019semantic} employ multi-level indoor semantic maps (rooms/obstacles/dynamics) to guide an RRT with local smoothing; 
Ryu \etal~\cite{ryu2020hierarchical} searches over a high-level graph before a local RRT smoothing step. Seder \etal~\cite{seder2011hierarchical}  plan over a hierarchical H-Graph of indoor regions with bridge nodes at doors, and refine paths on local occupancy grids before execution. However, these methods do not utilize the heirarchical information to limit and partition the sampling space, which highly limits scalability in large, complex environments and complicates replanning. 
\color{black}

\vspace{-4pt}
\subsection{3D Scene Graphs for Path Planning}

The structured representation of 3D scene graphs has been predominantly employed for task planning by using their hierarchical structure~\cite{agia2022taskography,ray2024task}, and even incorporating natural language processing~\cite{rana2023sayplan,gu2024conceptgraphs} by using \ac{LLM}s to match keywords in human-intuitive commands to nodes in the scene graph. 
Other works have utilized scene graphs to reason~\cite{amiri2022reasoning}, infer~\cite{rajvanshi2024saynav}, or learn~\cite{ravichandran2022hierarchical} navigation actions and policies. 
%However, their application to low-level path planning in geometric spaces remains highly underexplored.
\color{ black}However, these approaches plan at the semantic level and then delegate motion to geometric planners that plan over a large area, without using semantics to constrain and partition the search space.\color{black}

Recently, HOV-SG~\cite{Werby-RSS-24} processes natural language queries across three levels of abstraction (floor, room, and object) to exploit the hierarchical structure of a scene graph and progressively narrow down the solution space. However, this is only used to identify the goal node for navigation; the path planning still relies on a Voronoi graph over the entirety of the free space, which is not scalable to large areas.

The approach closest to ours is that of Ray \etal~\cite{ray2024task}, who proposed a task and motion planning framework over 3D scene graphs. 
Their method partitions the traversable space into \textit{places} and employs a tri-level planner for hierarchical planning. 
However, the \textit{places} are structured in a grid-like fashion and lack high-level semantic meaning. Moreover, their focus is primarily on plan feasibility, without exploiting the potential gains in efficiency or parallelism that S-Path offers.

\vspace{-4pt}
\color{ black}\subsection{Replanning in Motion Planning}
Another line of work accelerates planning and replanning by reusing geometric search effort. AIT*/EIT* \cite{strub2022adaptively} exploit problem-specific cost/effort heuristics, yielding in speed-ups; while EIRM* \cite{hartmann2022effort} extends this reuse to the multi-query setting. Further, E-Graphs \cite{phillips2012graphs} bias search along trajectories extracted from prior solutions to avoid re-exploration in configuration space. However, these approaches do not exploit semantic abstractions and scene graph topology to partition the search and localize replanning; thereby limiting scalability. \color{black}

% ======================================================================================================
% \section{System Overview}
\section{Methodology}
\begin{figure}[t]
    \vspace{2pt}
    \centering
    \includegraphics[width=0.97\columnwidth]{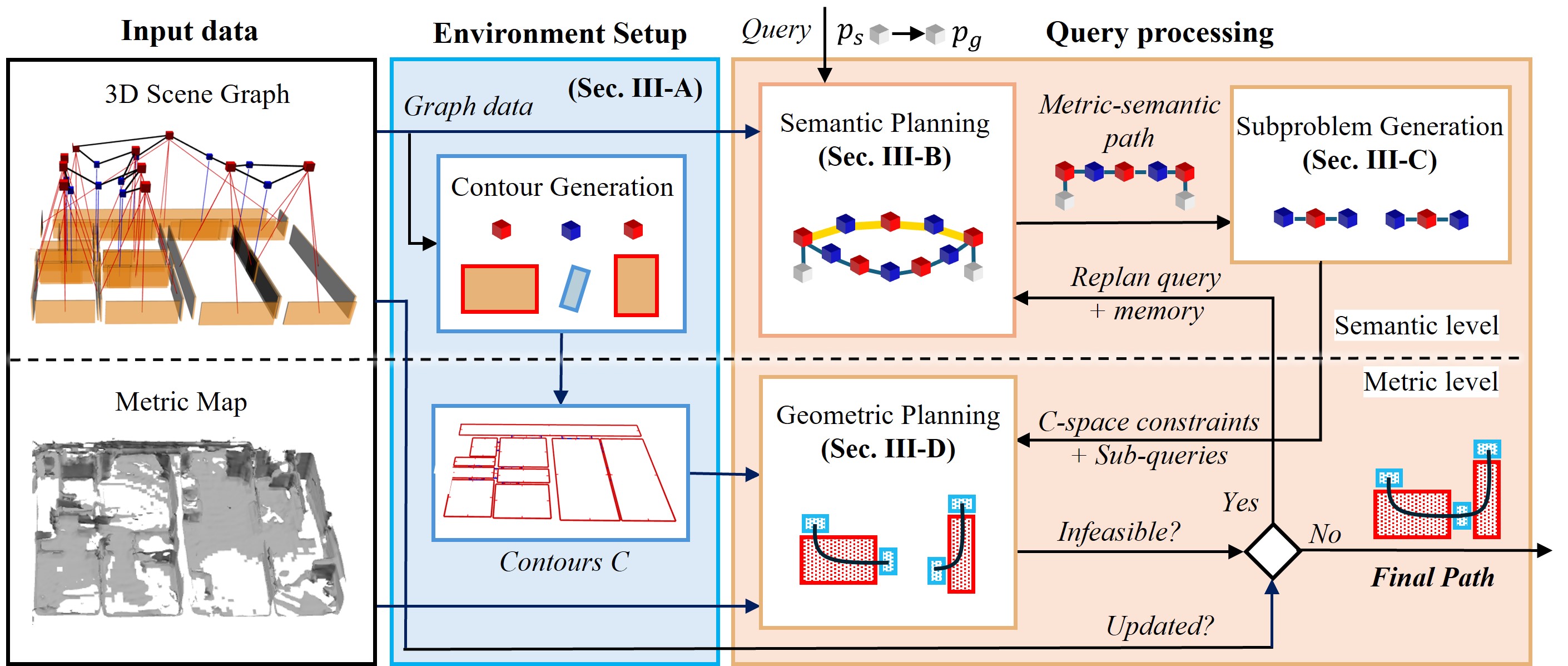}
    \caption{
    S-Path architecture: A 3D scene graph and metric map are used as input to set up the environment, which consists of contours. A start-to-goal query is decomposed into subproblems that constrain the C-space for faster geometric planning. Solutions to these are joined to form the final path. If the path is infeasible \textcolor{ black}{or the scene graph is updated}, reusing results from subproblems solved during the initial attempt speeds up replanning.
    \
    }
    \label{fig:system-overview}
    \vspace{-5pt}
\end{figure}

The architecture of S-Path is outlined in \Cref{fig:system-overview}.  
As \textit{Input Data} (\Cref{sec:input_data}), \spath requires a metric-semantic indoor 3D scene graph that includes information about rooms and doorways, and a metric mesh of the environment.  
Provided this input, a one-time \textit{Environment Setup} (\Cref{sec:environment_setup}) generates contours that associate nodes in the scene graph with corresponding regions in the metric mesh.
Subsequently, for each path planning query, \textit{Semantic Planning} (\Cref{sec:semantic-search}) is performed on the scene graph to generate a semantic path, restricting the problem to only the relevant areas of the environment.  
This problem is further decomposed through \textit{Subproblem Generation} (\Cref{sec:subproblem-solver}), which breaks the task into smaller, independent doorway-to-doorway subproblems.  
These subproblems are solved in parallel during \textit{Geometric Planning} (\Cref{sec:geometric-search}) using sampling-based planners.  
In the event that the path is infeasible, \textit{Replanning} (\Cref{sec:replanning}) ensures an efficient reroute.

% \subsection{3D Scene Graph}
\vspace{-3pt}
\subsection{Input Data}
\label{sec:input_data}
S-Path requires both an indoor 3D metric-semantic scene graph to support semantic planning, and a metric mesh of the environment for geometric planning.

The 3D scene graph is an undirected topological graph that encodes both metric-semantic information and relational data from an indoor environment.
S-Path expects the graph to contain at least the following types of nodes: 
\begin{itemize}
    \item \textit{Walls}, where each wall \(\mathbf{W}_i = \{\boldsymbol{\pi}_i ~|~ \boldsymbol{\pi}_i \in \mathbb{R}^4\}\) is defined by the plane equation \(\boldsymbol{\pi}_i = [\mathbf{n}_i^\top, d_i]^\top\), with the normal \(\mathbf{n}_i \in \mathbb{R}^3\) pointing inside the room and scalar offset \(d_i\).
    \item \textit{Rooms}, where each room \(\mathbf{R}_i = \{\mathbf{\Psi}_i, \boldsymbol{\rho}_i ~|~ \boldsymbol{\rho}_i \in \mathbb{R}^3\}\) is defined by its centroid \(\boldsymbol{\rho}_i\) and a set of walls bounding it, i.e., \(\mathbf{\Psi}_i = \{\mathbf{W}_{i,1}, \mathbf{W}_{i,2}, \,\hdots, \mathbf{W}_{i,m} \}\) with \(m \geq 3\).
    
    \item \textit{Doorways}, where each doorway \(\mathbf{D}_i = \{\boldsymbol{\delta}_i, w_i, s_i ~|~ \boldsymbol{\delta}_i \in \mathbb{R}^3, w_i \in \mathbb{R}, s_i \in \{0,1\} \}\) is defined by its centroid \(\boldsymbol{\delta}_i\), width \(w_i\), and a binary state \(s_i\) indicating whether the doorway is traversable.
\end{itemize}

This graph can be generated from real-time LiDAR  \cite{sgraphsp} or visual \cite{hydra} measurements or extracted from real or synthetic environment maps using \ac{BIM} \cite{shaheer2023graph}, or from public datasets \cite{floordataset}.

The metric mesh is obtained from a voxel-based map, 
\color{ black}
with obstacles in the environment represented as occupied voxels,
\color{black}
which is converted to a \ac{ESDF}, using Voxblox \cite{Oleynikova2017a}.
This forms the foundation for path planning heuristics and collision checking.

The aforementioned input data is used once to set up the environment and then can be updated in response to changes.
For path planning, a query with start and goal points, $\mathbf{p}_s$ and $\mathbf{p}_g$, is required. 
This query can also be provided in a human-interpretable form, i.e., from a start room $\mathbf{R}_s$ to an end room $\mathbf{R}_g$, in which case the centroids of the rooms are used: $\mathbf{p}_s=\boldsymbol{\rho}_s$ and $\mathbf{p}_g=\boldsymbol{\rho}_g$.
Moreover, a planning time budget, denoted by \textit{time-to-plan} (\(ttp\)), needs to be specified.

\vspace{-3pt}
\subsection{Environment Setup}
\label{sec:environment_setup}
    
\textbf{Semantic graph generation }
The scene graph is transformed into an undirected, weighted \textit{semantic graph} that connects the semantic elements (rooms and doors) within the scene, which is used to compute a semantic path.
Each connection between nodes is assigned a cost equal to the Euclidean distance between the centroids of the elements representing those nodes.
Lastly, edges connected to non-traversable doorways are assigned an infinite weight.

\begin{figure}[t]
    \vspace{2pt}
    \centering
    \includegraphics[width=0.75\columnwidth]{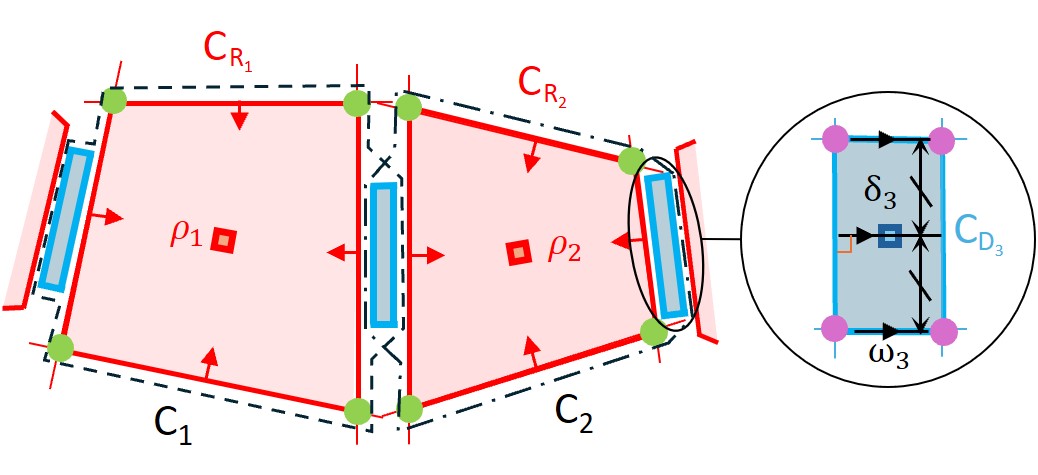}
    \caption{
    Room contours $\mathbf{C}_{\mathbf{R}_i}$ are derived from the intersection points of wall lines.
    Doorway contours $\mathbf{C}_{\mathbf{D}_i}$ are fitted between two rooms using the doorway's center, its width, and the closest room segment as a reference.
    $\mathbf{C}_i$ contours are the union of those of rooms and doorways.
    }
    \label{fig:contours}
    \vspace{-5pt}
\end{figure}

\smallskip
\textbf{Contour generation }
The wall planes information in the scene graph is used to generate room and doorway \textit{contours} \(\mathbf{C} = \{\mathbf{C}_R,\mathbf{C}_D\}\).
% that define their respective boundaries.
Contours serve two purposes: first, they restrict sampling-based geometric planners to areas that contribute to the final path; second, they enable the association of geometric locations with semantic elements, such as rooms and doorways, via a \textit{point-in-polygon} test.
Room polygons are derived by projecting wall planes into 2D lines, computing all possible intersections, and generating line segments between them (see \Cref{fig:contours})
These segments are then connected at their endpoints to form enclosed 2D polygons.
% Inspired by \cite{Ferreira2023}, 
% The final room contours are obtained by extruding these 2D polygons along the \textit{z}-axis to match the room's height.
Doorway polygons are constructed by casting a ray through the centroid of the doorway, perpendicular to the closest wall segment of the closest room. 
Two additional rays are cast parallel to this ray at a half-width offset on either side (refer to \Cref{fig:contours}). 
The four resulting intersection points with the walls of the connecting rooms are used to construct the doorway polygon.
Both the room and doorway polygons are extruded along the \textit{z}-axis to match the respective room height, \textcolor{ black}{forming 2.5D contours.}
% , resulting in a 2.5D contour map.

It is important to note that the quality of contour generation is dependent on the accuracy of the wall information associated with each room in the input scene graph.  
Moreover, the proposed algorithm does not work for concave-shaped rooms; however, such rooms are uncommon in indoor environments and, if present, can either be split into convex rooms connected by \textit{artificial} doorways or be represented by a convex polygon (rectangle in the simplest case) that spans the area of the room. 

% ======================================================================================================
% % ======================================================================================================
\vspace{-4pt}
\subsection{Semantic Planning}
\label{sec:semantic-search}
% To account for this, the edges in the semantic graph corresponding to the blocked doorways are removed.  
Given the input query $\mathbf{p}_s \rightarrow \mathbf{p}_g$, \textit{point-in-polygon} tests are performed over room contours to determine the rooms associated with the start and goal points, denoted as $\mathbf{R}_s$ and $\mathbf{R}_g$, respectively.  
The start and goal points may also lie within doorway contours; however, since the area of doorway contours is typically much smaller than that of room contours, and the overall process remains similar, we assume, without loss of generality, that these points lie within room contours.
Once the relevant contours are identified, an $A^*$ search is then executed on the semantic graph to compute the shortest semantic path from $\mathbf{R}_s$ to $\mathbf{R}_g$, as follows:
\begin{align}
    \begin{split}
       \mathbf{\Pi}_\text{sem} := \mathbf{R}_s \rightarrow \mathbf{D}_k \rightarrow \mathbf{R}_{r} \rightarrow \cdots \rightarrow \mathbf{D}_{k+n} \rightarrow \mathbf{R}_g
       \label{eq:semantic-path}
    \end{split}
\end{align}
\noindent where $\mathbf{D}_k$ and $\mathbf{D}_{k+n}$ is the first and last doorway in the semantic path with $n+1$ doorways and $n+2$ rooms.
Furthermore, since the semantic graph is generally sparse, performing an $A^*$ search on it incurs negligible computational cost compared to a full search on the metric layer.  
\(\mathbf{\Pi}_{\text{sem}}\) is used to generate coarse waypoints for the robot to follow, in the form of the center points of the doorways along the path. 
These waypoints define the coarse geometric path, \(\mathbf{\Pi}_{\text{geo}}\), given by:
\begin{align}
    \begin{split}
       \mathbf{\Pi}_\text{geo} := \mathbf{p}_s \rightarrow \boldsymbol{\delta}_k \rightarrow \boldsymbol{\delta}_{k+1} \rightarrow \cdots \rightarrow \boldsymbol{\delta}_{k+n} \rightarrow \mathbf{p}_g, \\
       \textit{via:}\hspace{0.1cm} \mathbf{C}^{\prime} = \mathbf{C}_{\mathbf{R}_s} \cup \bigcup\limits_{i=0}^{n-1} \mathbf{C}_{\mathbf{R}_{r+i}} \cup \bigcup\limits_{i=0}^{n} \mathbf{C}_{\mathbf{D}_{k+i}} \cup \mathbf{C}_{\mathbf{R}_g}
       \label{eq:semantic-search}
    \end{split}
\end{align}
\noindent where the reduced sample space $\mathbf{C}^{\prime}$ is composed of the room and doorways contours within the semantic path and \(\mathbf{\delta}_k\) is the center point of $\mathbf{D}_k$.
It is important to note that introducing intermediate waypoints forces the robot to pass exactly through the center of each doorway, which may adversely affect path optimality compared to solving the global planning problem directly from \(\mathbf{p}_s \rightarrow \mathbf{p}_g\). However, in practical scenarios, doorways are often narrow, meaning their center points are typically close to the globally optimal crossing points, thus minimizing the impact of subproblem decomposition on the overall path length.
Another consideration is that the semantic graph is constructed with edge weights corresponding to the Euclidean distance between room centers and the centers of the connecting doorways. 
If room sizes are highly uneven, for instance, a very large room that is only partially traversed, the resulting semantic path may be suboptimal. 
Nonetheless, these added costs are arguably outweighed by the planning-time benefits that S-Path achieves through decomposing the problem, as validated in \Cref{sec:evaluation}.

% ======================================================================================================
\vspace{-2pt}
\subsection{Subproblem Generation}
\label{sec:subproblem-solver}

The coarse geometric path in \Cref{eq:semantic-search} can be decomposed into independent subproblems
% , each of which presents a smaller task for geometric planning,
as follows: 
% These subproblems are also independent in nature, enabling parallel computation using a thread pool. 
\begin{equation}
\begin{alignedat}{3}
   1&: \mathbf{p}_s &&\rightarrow \boldsymbol{\delta}_k, &&\phantom{0}\textbf{via: } \mathbf{C}_{1} \\
   2&: \boldsymbol{\delta}_k &&\rightarrow \boldsymbol{\delta}_{k+1}, &&\phantom{0}\textbf{via: } \mathbf{C}_{2} \\
   & &&\cdots &&\\
   % i&: \boldsymbol{\delta}_{k+i-2} &&\rightarrow \boldsymbol{\delta}_{k+i-1}, &&\phantom{0}\textbf{via: } \mathbf{C}_{i}\\
   % & &&\cdots &&\\
   n+2&: \boldsymbol{\delta}_{k+n} &&\rightarrow \mathbf{p}_g, &&\phantom{0}\textbf{via: } \mathbf{C}_{n+2}
\end{alignedat}
\label{eq:subproblems}
\end{equation}
\noindent where for the $i^{th}$ subproblem, $\mathbf{C}_i$ refers to the composed contours, \ie the boolean union of the room and doorway contours required to solve the sub-problem.
The subproblems vary in size; hence, time allocation during geometric planning must account for this to ensure a fair distribution of time.
Therefore, for each subproblem, an effort heuristic is computed, correlating with the size of the subproblem as follows:
\begin{align}
 e_i = \begin{cases} 
      \norm{\boldsymbol{\delta}_{k} - \mathbf{p}_s} + \sqrt{a_1} & i = 1 \\
      \norm{\boldsymbol{\delta}_{k+i-2} - \boldsymbol{\delta}_{k+i-1}} + \sqrt{a_i} & 2\leq i\leq n+1 \\
      \norm{\mathbf{p}_g - \boldsymbol{\delta}_{k+n}} + \sqrt{a_{n+2}} & i = n+2
   \end{cases}
    \label{eq:effort}
\end{align}
\noindent where $a_i$ is the area of the composed contour $\mathbf{C}_i$.
Subproblems with higher heuristic values receive more computational resources and time.
Moreover, to prevent excessively small time allocations, subproblems with heuristic values less than a predefined value, defined as a proportion of an equal split, are merged with one of their neighboring subproblems, specifically, the neighbor with the lower effort heuristic, among the at most two adjacent candidates.
Following this, the effort heuristics in \Cref{eq:effort} are recomputed.
This ensures that planning does not fail due to avoidable time constraints and also relaxes the strict requirement of passing exactly through the center of doorways for the merged subproblems.  
This merging process is performed iteratively until no subproblem remains excessively small. 
\color{ black}
Additionally, we note that although the composed contours $\mathbf{C}_i$ are 2.5D, the resulting subproblems correspond to decompositions of the underlying 3D mesh and are therefore defined entirely in 3D.
\color{black}
% , after which the final set of subproblems is ready for geometric planning.
% ======================================================================================================
\begin{figure}[t]
% \centering
\vspace{5pt}
\begin{subfigure}{0.159\textwidth}
 \centering
\includegraphics[width=1.0\textwidth]{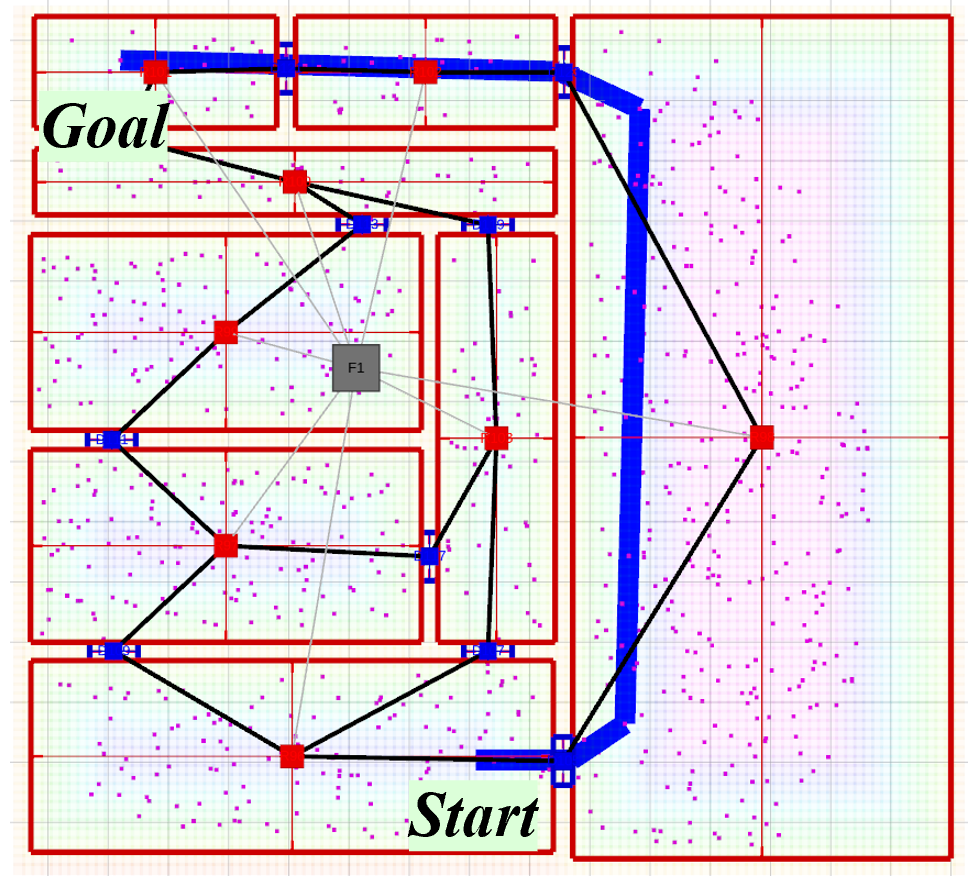}
% \caption{$(o)$}
\caption{}
\label{fig:baseline}
\end{subfigure}
\begin{subfigure}{0.159\textwidth}
% \centering
\includegraphics[width=1.0\textwidth]{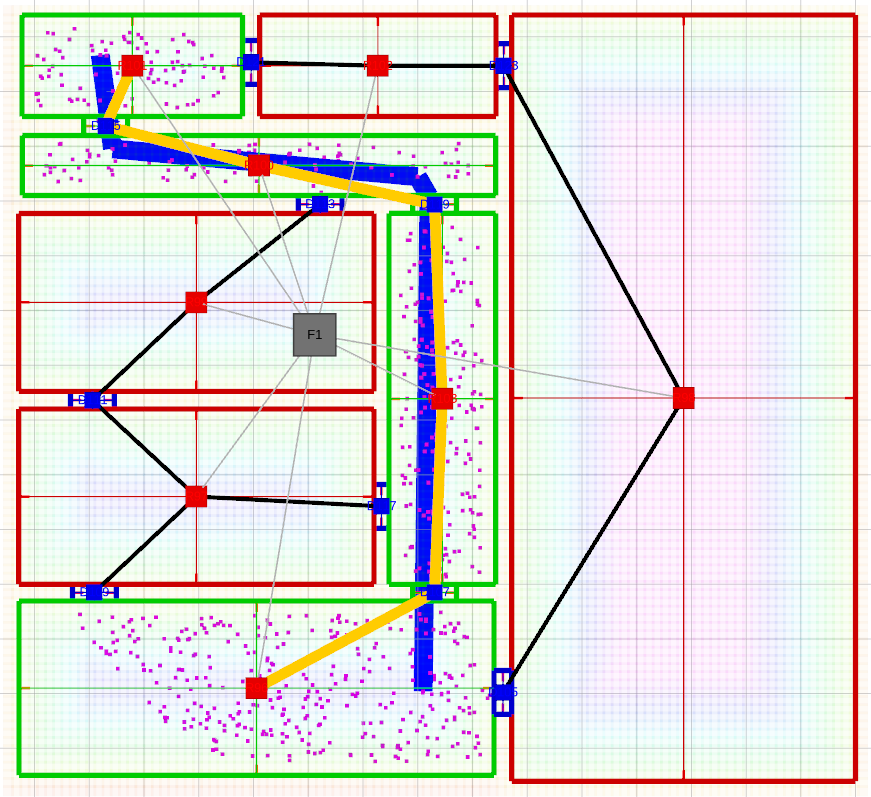}
% \caption{$(o+r)$}
\caption{}
\label{fig:reduced_space}
\end{subfigure}
\begin{subfigure}{0.159\textwidth}
% \centering
\includegraphics[width=1.0\textwidth]{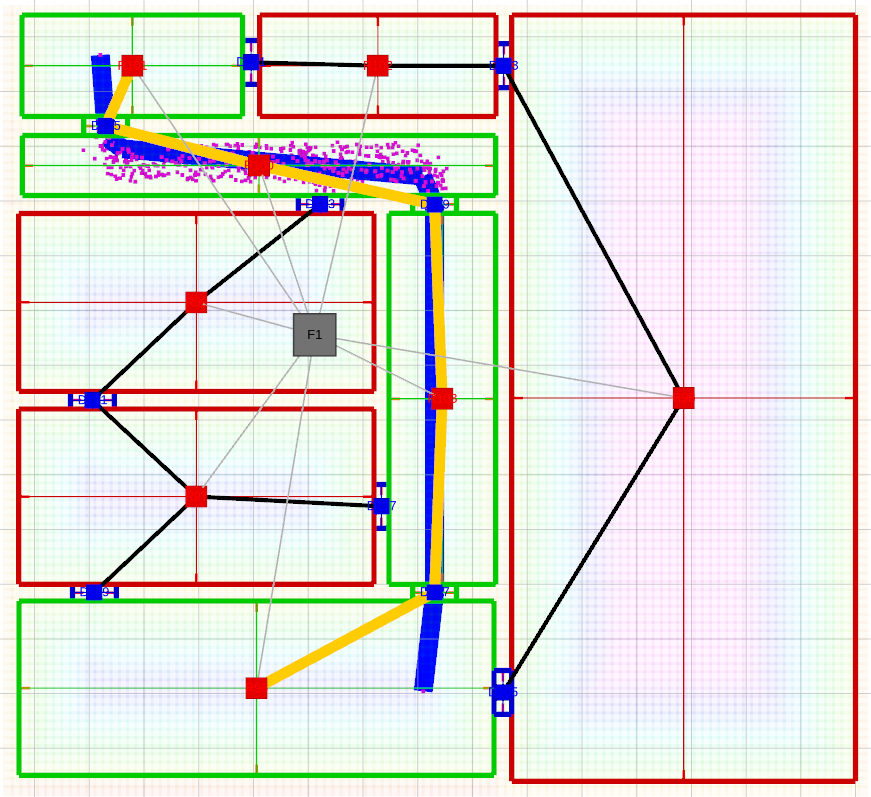}
% \caption{$(o+r+s)$}
\caption{}
\label{fig:spath}
\end{subfigure}
\caption{Illustration of the features of \spath: (a) The baseline creates many samples that do not contribute to the final path, thus finding a suboptimal solution given a limited time. (b) Restricting sample space to $\mathbf{C}^{\prime}$ leads to denser coverage of only relevant regions (c) Subproblem decomposition divides results in smaller and sometimes trivial (notice the straight line path in the last two rooms) problems.}
\label{fig:snav-planning}
\vspace{-5pt}
\end{figure}

\vspace{-4pt}
\subsection{Geometric Planning}
\label{sec:geometric-search}
The geometric planning in \spath employs  
a thread pool approach, distributing subproblems across \(m\) parallel workers executing planners from the \ac{OMPL} library \cite{Sucan2012}. 
\color{ black}
The robot is modeled as a sphere with a radius equal to its largest dimension plus a safety margin. 
Collision checking is performed against the metric mesh represented by the ESDF  introduced in Section \ref{sec:input_data}. 
Moreover, the planner assumes a holonomic robot and models the C-space as consisting of the 3D positional dimensions ($x$, $y$, and $z$).
\color{black}

Since the quality of the path produced by sampling-based planners is directly related to the sampling time (a denser set of samples is more likely to yield shorter paths), the \(ttp\) for each subproblem is distributed proportionally to the effort heuristics defined in \Cref{eq:effort}.
% In addition to being solved in parallel, the local subproblems benefit from a reduced sample space, as the composed contours are used to restrict the underlying mesh. 
Planning for smaller subproblems avoids unnecessary collision checks and rewiring of vertices that do not contribute to the local solution, leading to significantly faster planning. 
This is evident in \Cref{fig:snav-planning}, where, compared to planning over the global problem (see \Cref{fig:baseline}), restricting the C-space to \(\mathbf{C}'\) results in denser coverage and a shorter path (see \Cref{fig:reduced_space})
Further subproblem decomposition can even yield point-to-point, obstacle-free subproblems, which are trivial to solve using an informed planner like BIT*,
as demonstrated by the empty regions of the first two and last rooms along the path in \Cref{fig:spath}, where no sampling was needed.
Once all subproblems have been solved, local solutions are stitched at the respective doorway centers to get the final path. 
\color{ black}
If the geometric planner fails to compute a valid path due to a mismatch between the actual traversability and the one assumed by the semantic graph, the replanning mechanism is triggered, as described in the following subsection.
\color{black}

\vspace{-4pt} 
\subsection{Replanning}
\label{sec:replanning}
\color{ black}
Changes are monitored at both the semantic and geometric levels. When the semantic graph is affected, for example, due to closed doorways or newly introduced obstructions, or when the geometric planner fails to compute a feasible path, our efficient replanning mechanism is invoked. 
% In such cases, S-Path first updates the environment representation and reconstructs the semantic graph, assigning infinite edge weights to affected edges or edges connected to inaccessible nodes.
In such cases, S-Path reconstructs the semantic graph, assigning infinite edge weights to affected edges or edges connected to inaccessible nodes.
Then, edges corresponding to already planned portions of the path are updated with reduced weights (scaling by a factor less than 1), encouraging their reuse during subsequent replanning.
Semantic planning is then re-executed and previously solved subproblems are retrieved from memory to prevent redundant computation.
% starting from the node on the original path closest to the detected blockage. 
% Previously solved subproblems are retrieved from memory to prevent redundant computation, and only 
Only the updated portions of the semantic path are considered during geometric planning, thereby reducing replanning time.
\color{black}
Finally, the newly computed and cached path segments are joined to form the final path.

% ======================================================================================================
\section{Evaluation}
\label{sec:evaluation}
\subsection{Validation Methodology}
To the best of our knowledge, there is no prior work directly comparable to ours. 
The most closely related approach~\cite{ray2024task} focuses on planning feasibility instead of efficiency, and is not open-source. 
Therefore, we evaluate performance improvements of \spath relative to the underlying sampling-based planners. 
Three planners were selected:
1) BIT* \cite{Gammell2015} as a tree-style planner with built-in heuristics.
2) PRM* \cite{prmstar} as an unbiased coverage-of-space planner; and
3) RRT* \cite{Karaman2011} as a tree-style planner with no built-in heuristics.
The validation is divided into two parts: sequential and parallel execution. 
The sequential evaluation ablates the effects of the individual components of S-Path, while the parallel evaluation showcases its scalability.

\smallskip
\begin{figure}
    \vspace{5pt}
    \begin{subfigure}{\linewidth}
        % Note: Rviz pitch angle: 0.6
        \centering
        \resizebox{0.8\linewidth}{!}{\includegraphics{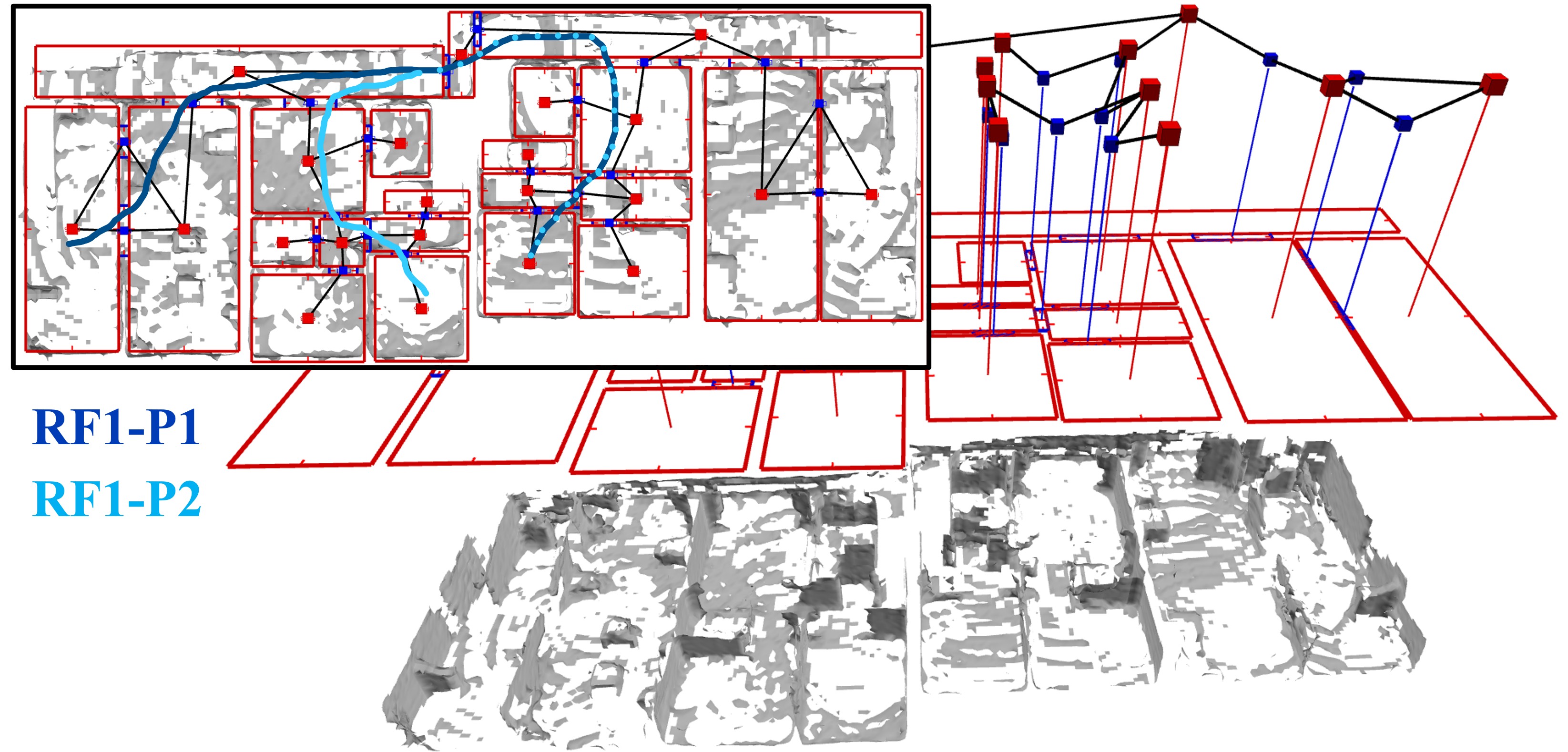}}
        % \resizebox{0.8\linewidth}{!}{\includesvg{pictures/datasetRF1.svg}}
        \caption{RF1: Real construction site, augmented by the scene graph built from its BIM model.}
        \label{fig:rf1}
    \end{subfigure}
    \begin{subfigure}{\linewidth}
        % Note: Rviz pitch angle: 0.6
        \centering
        \resizebox{0.8\linewidth}{!}{\includegraphics{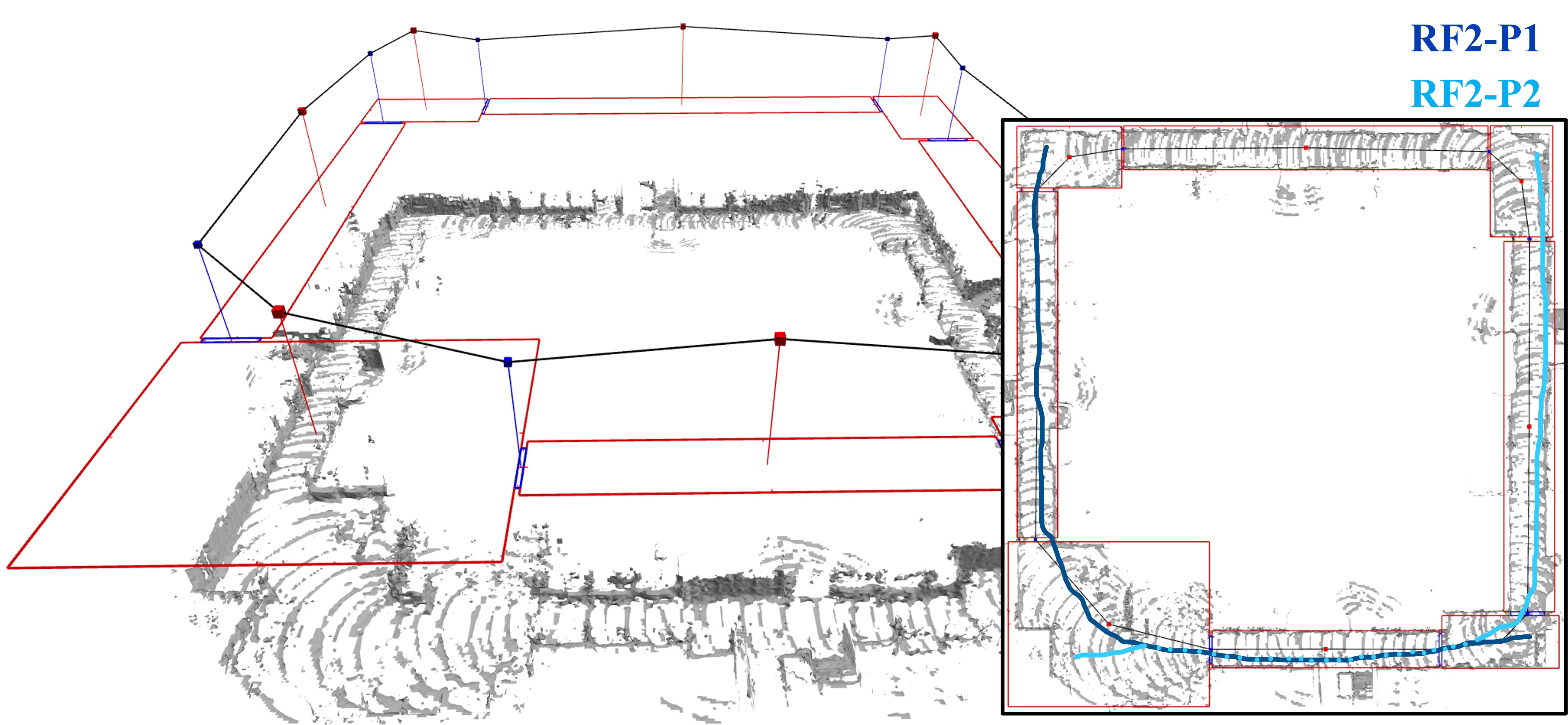}}
        % \resizebox{0.8\linewidth}{!}{\includesvg{pictures/datasetRF2.svg}}
        \caption{RF2: Campus building augmented by the scene graph built from its BIM model, featuring long, wide corridors with some obstacles.}
        \label{fig:rf2}
    \end{subfigure}
    \begin{subfigure}{\linewidth}
        \centering
        \resizebox{0.8\linewidth}{!}{\includegraphics{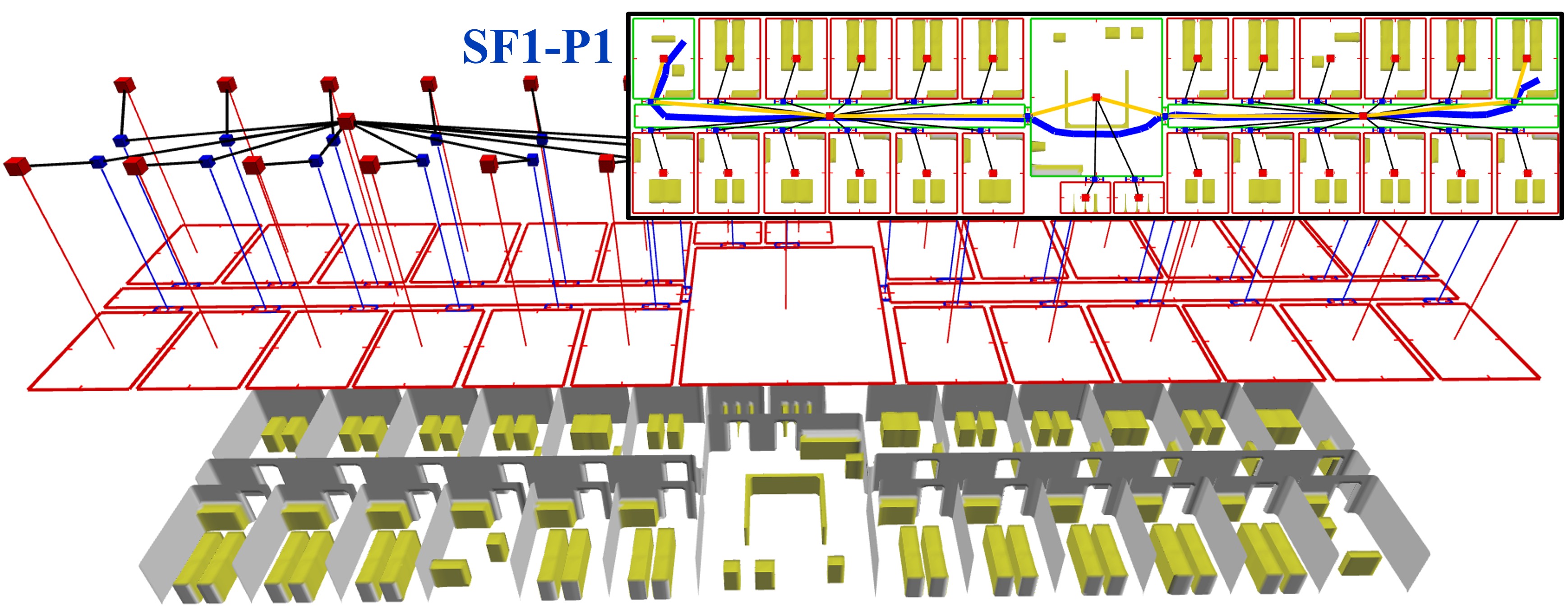}}
        % \resizebox{0.8\linewidth}{!}{\includesvg{pictures/datasetSF1.svg}}
        \caption{SF1: Synthetic office floor environment with several obstacles. The offices on either side of the floor are connected via a central corridor that leads to a shared space in the center.}
        \label{fig:sf1}
    \end{subfigure}
    \begin{subfigure}{\linewidth}
        \centering
        \resizebox{0.8\linewidth}{!}{\includegraphics{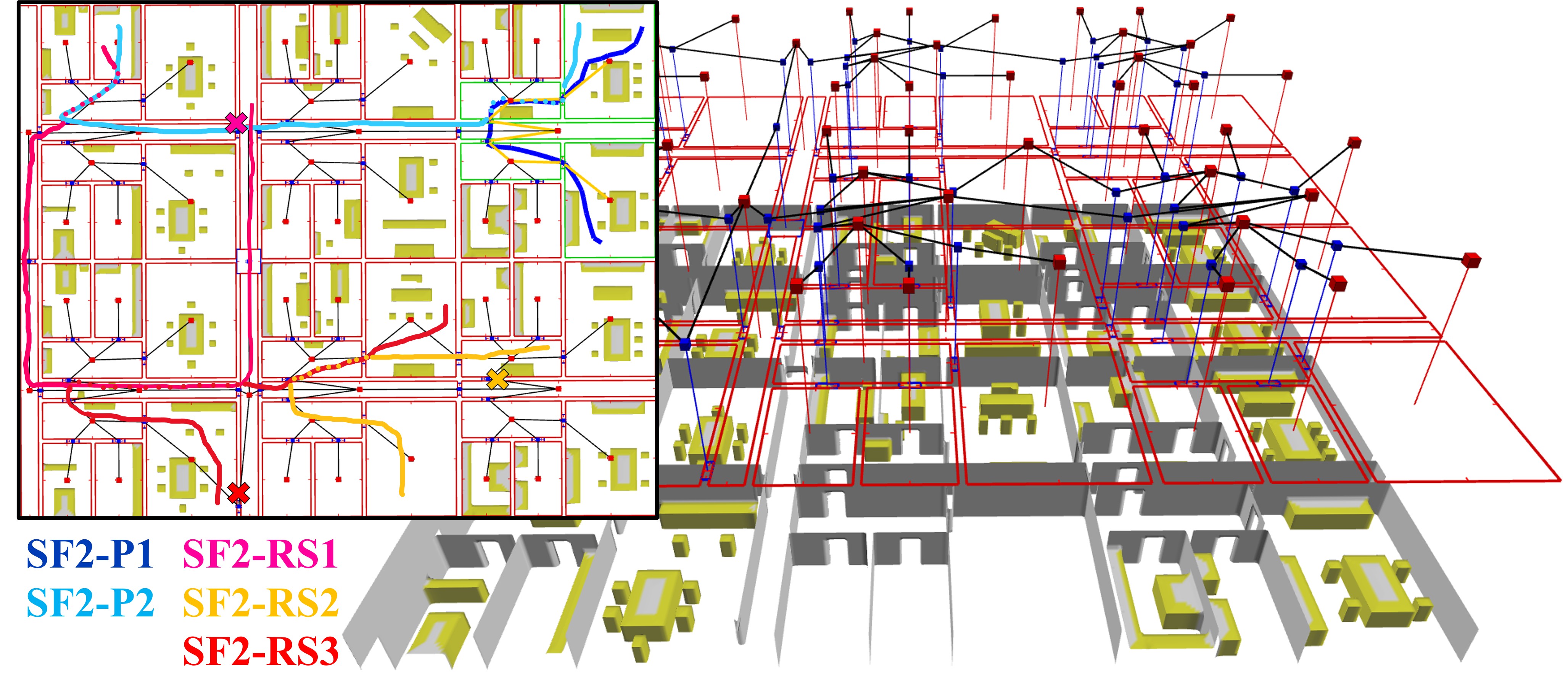}}
        % \resizebox{0.8\linewidth}{!}{\includesvg{pictures/datasetSF2.svg}}
        \caption{SF2: Synthetic environment representing a large apartment floor with several units and long, narrow corridors.}
        \label{fig:sf2}
    \end{subfigure}
    \caption{Real and synthetic environments used for evaluations. The paths shown in the top-down orthogonal view are the specific queries used from each environment in the sequential execution evaluation. -P* represents a query for planning scenarios and -RS* for replanning scenarios.
    % Real maps are built from recorded LiDAR data and augmented with architectural plans featuring rooms and doorways.
    }
    \label{fig:datasets}
    \vspace{-5pt}
\end{figure}

\noindent \textbf{Environments} 
Two real and two synthetic environments \textcolor{ black}{(up to $2900\,m^2$ in area)} are used for evaluations and shown in \Cref{fig:datasets}.
For real environments (\Cref{fig:rf1,fig:rf2}), LiDAR scans are used to generate the scene graphs using \cite{shaheer2023graph} and manually constructed BIM models, and the metric mesh following \cite{Oleynikova2017a}.
% \ac{ESDF} maps to be used as the metric meshes for geomtric planning.
% The scene graphs are generated using \cite{shaheer2023graph} by overlaying BIM models manually on a top-down map of the environment.
Moreover, concave rooms are represented by a spanning rectangular contour as described in \Cref{sec:environment_setup} (see bottom-left of environment RF2).
Furthermore, since the real environments are limited in size and complexity, we employ synthetically generated environments (\Cref{fig:sf1,fig:sf2}) that cover relatively larger areas with multiple bounded obstacles. 
In this case, the scene graph is derived from the positions of rooms and doorways in the synthetic environment, while the voxels representing obstacles and walls are used to generate an occupancy grid that is converted to an \ac{ESDF}, using \cite{Oleynikova2017a}.

\smallskip
\noindent \textbf{Sequential execution methodology }  
A version of S-Path that solves all subproblems in sequence, referred to as S-Path(s), is compared against the aforementioned planners operating on the global problem, which is denoted by \textbf{I} (refer to \Cref{fig:baseline}).
In addition, other ablations are also evaluated, specifically, \textbf{II} (refer to \Cref{fig:reduced_space}), which extends \textbf{I} by adding C-space restrictions following \Cref{eq:semantic-search}, and \textbf{III} (refer to \Cref{fig:spath}), which extends \textbf{II} by adding subproblem decomposition following \Cref{eq:subproblems}, but without replanning memory.

Performance is evaluated on 10 queries divided into two scenarios: planning and replanning.
For the planning scenarios, four queries are used from the synthetic environments and four from the real environments. 
These queries differ in path length and complexity as can be seen in \Cref{fig:datasets}. 
$ttp_{min}$ is set to $1\,\mathrm{ms}$ while $ttp_{\max}$ is set to $6\,\mathrm{s}$.
Moreover, in these scenarios, the ablation \textbf{III} and S-Path(s) are equivalent, as there is no replanning involved, so \textbf{III} is omitted from the results.
For the replanning scenarios, only the large and complex SF2 environment is used, as it offers multiple routes between most node pairs. 
This enables the intentional blocking of key doorways to create challenging evaluation cases. 
In these scenarios, when the planner receives an update to the scene graph that renders the current path infeasible due to an untraversable doorway, it updates the start position to the closest doorway along the path preceding the obstruction.
All ablations are then re-executed for the modified query, incorporating the necessary updates to the semantic graph and mesh owing to the blockage. 
Additionally, S-Path(s) retains memory of the previously computed plan to inform subsequent replanning.
The replanning scenarios are generally more difficult than planning scenarios, as heuristics for sampling-based solvers are based on Euclidean distance, which might fail in the presence of a closed doorway because the shortest path could require moving in the opposite direction of the goal initially.
To accommodate this, $ttp_{\min}$ is set to $25\,\mathrm{ms}$ and $ttp_{\max}$ is increased to $10\,\mathrm{s}$.
Overall, three replanning scenarios are evaluated as visualized in \Cref{fig:sf2}.

All ablations \(\mathcal{A} = \{\textbf{I}, \textbf{II}, \textbf{III}, \text{S-Path(s)}\}\) 
% are evaluated with respect to their computational cost ($ttp$) and path length.
are evaluated using an
% To this end, an overall
efficiency metric \(\bar{\eta}\), which considers both $ttp$ and path length, while also accounting for robustness by repeating each computation 100 times to obtain results at a $\SI{95}{\%}$ success rate.
In this regard, computational efficiency \(\eta_{ttp, i}\) measures the improvement in time-to-plan (\(ttp\)) relative to ablation \textbf{I}:
\begin{align}
    \eta_{ttp, i} = \frac{ttp_{95\%, \textbf{I}}}{ttp_{95\%, i}}, \hspace{0.2cm} \forall\; i \in \mathcal{A}
    \label{eq:efficiency-ttp}
\end{align}
\noindent where a higher $\eta_{ttp, i}$ indicates better computational performance relative to the underlying sampling-based planner.
The $ttp_{95\%}$ is computed by first defining bounds, i.e., $ttp \in [ttp_{\min}, ttp_{\max}]$, and then sampling values logarithmically within this range to obtain a series of candidate $ttp$ values. 
For each sampled value, the success rate is calculated as the percentage of runs that result in a feasible path. 
Finally, $ttp_{95\%}$ is estimated using piecewise linear interpolation over the success rates to determine the minimum $ttp$ at which at least $95\%$ of the runs are successful.
\color{ black}
It should be noted that $ttp$ also includes the time for subproblem decomposition and for stitching local solutions into a final path, wherever applicable.
\color{black}
% This metric emphasizes how efficiently the method uses computational resources while maintaining a high success rate.
Similarly, the path efficiency \(\eta_{l, i}\) is defined as the ratio of the shortest path length \(l_{conv, i}\) to the path length at the \(\SI{95}{\%}\) success mark \(l_{95\%, i}\), determined similarly to \(ttp_{95\%}\) by running 100 trials for each ablation:
\begin{align}
    \eta_{l, i} = \frac{l_{conv, i}}{l_{95\%, i}}, \hspace{0.2cm} \forall\; i \in \mathcal{A}
    \label{eq:efficiency-path-length}
\end{align}
\noindent which favors consistent path lengths and is used as a measure of robustness.
Finally, the combined efficiency $\eta_i$ of a planning method is given by:
\begin{align}
    \eta_i = \eta_{l, i} \; \eta_{ttp, i}, \hspace{0.2cm} \forall\; i \in \mathcal{A}
    \label{eq:efficiency}
\end{align}
\noindent leading to the normalized efficiency gain $\bar{\eta}_i$ as:
\begin{align}
    \bar{\eta}_i = \frac{\eta_i}{\eta_{\textbf{I}}}, \hspace{0.2cm} \forall\; i \in \mathcal{A}
    \label{eq:efficiency-normalized}
\end{align}
\noindent which is considered as the final unified metric for comparisons.

\sisetup{group-separator = {}}
\begin{table*}
    \vspace{3pt}
    \caption{Efficiency gains of the tested ablations across all queries using \textit{PRM*}, \textit{BIT*}, and \textit{RRT*} planners. The best and second-best efficiency results for each configuration are boldfaced and underlined respectively. While S-Path(s) shows improvement in the \textit{time-to-plan}, $ttp$, for all configurations, path optimality varies. Therefore, path length results by S-Path(s) are highlighted in \colorbox{greenl}{green} or in \colorbox{red}{red} if they are more than $2\%$ shorter or longer compared to \textbf{I}, respectively; and in \colorbox{yellow}{yellow} otherwise. }
    \centering
    \scriptsize
    \resizebox{0.96\linewidth}{!}{%
    \begin{tabular}{p{4pt}|c|l||S[table-format=-3.2, mode=text]cc|c||S[table-format=-3.2, mode=text]cc|c||S[table-format=-6.0, mode=text]cc|c}
        \toprule
            & \multicolumn{2}{c||}{} & \multicolumn{4}{c||}{\textbf{BIT*}} & \multicolumn{4}{c||}{\textbf{PRM*}} & \multicolumn{4}{c}{\textbf{RRT*}}\\
        % \midrule
        \cmidrule{2-15}
        % ====================================
        % RF1 P1
        % ====================================
            & \multicolumn{1}{c|}{\textbf{Scenario}} & \multicolumn{1}{c||}{\textbf{Ablation}}
            & $ttp_{\SI{95}{\%}}$ $[ms]$ & $l_{\SI{95}{\%}}$ $[m]$ & $l_{conv}$ $[m]$ & $\bar{\eta}$ 
            & $ttp_{\SI{95}{\%}}$ $[ms]$ & $l_{\SI{95}{\%}}$  $[m]$ & $l_{conv}$ $[m]$ & $\bar{\eta}$
            & $ttp_{\SI{95}{\%}}$ $[ms]$ & $l_{\SI{95}{\%}}$ $[m]$ & $l_{conv}$ $[m]$ & $\bar{\eta}$ \\
        \toprule
        \multirow{29}{*}{\rotatebox{90}{\textit{Planning Scenarios}}} &\multirow{3}{*}{\textbf{RF1-P1}} & \textbf{I} & 272.97  & 32.65 & 32.02 & 1.00             % BIT*
                 & 1537.47 & 32.51 & 32.10 & 1.00            % PRM*
                 & \textgreater 6000.00 & - & - & - \\        % RRT*
                 
        && \textbf{II} & 140.19 & 32.98 & 31.83 & \underline{1.92}             % BIT*
                 & 661.93 & 32.41 & 31.83 & \underline{2.39}              % PRM*
                 & \textgreater 6000.00 & - & - & - \\        % RRT*
                 
        && \textbf{S-Path(s)} & 104.77 & \cellcolor{red}{34.74} & \cellcolor{red}{34.28} & \textbf{2.62}         % BIT*
                 & 445.35 & \cellcolor{red}{34.52} & \cellcolor{red}{34.28} & \textbf{3.47}         % PRM*
                 & 1238.62 & 33.96 & 33.80 & - \\    % RRT*
        % \midrule
        \cmidrule{2-15}
        % ====================================
        % RF1 P2
        % ====================================
        &\multirow{3}{*}{\textbf{RF1-P2}} & \textbf{I} & 157.24  & 28.17 & 25.60 & 1.00             % BIT*
                 & 675.94 & 28.21 & 26.70 & 1.00            % PRM*
                 & \textgreater 6000.00 & - & - & - \\        % RRT*
                 
        && \textbf{II} & 86.40 & 29.10 & 26.96 & \underline{1.86}             % BIT*
                 & 279.80 & 28.98 & 27.31 & \underline{2.41}              % PRM*
                 & 3347.76 & 26.51 & 26.03 & - \\        % RRT*
                 
        && \textbf{S-Path(s)} & 34.23 & \cellcolor{red}{31.55} & \cellcolor{red}{30.91} & \textbf{4.95}         % BIT*
                 & 68.85 & \cellcolor{red}{31.96} & \cellcolor{red}{31.01} & \textbf{10.06}         % PRM*
                 & 742.91 & 30.80 & 30.66 & - \\    % RRT*
        % \midrule
        \cmidrule{2-15}
        % ====================================
        % RF2P1
        % ====================================
        &\multirow{3}{*}{\textbf{RF2-P1}} & \textbf{I} & 87.70  & 88.96 & 87.97 & 1.00             % BIT*
                         & 162.86 & 89.65 & 88.53 & 1.00            % PRM*
                         & 188.01 & 88.31 & 88.71 & \underline{1.00} \\        % RRT*
                         
                && \textbf{II} & 68.37 & 89.98 & 89.11 & \underline{1.28}             % BIT*
                         & 159.15 & 89.93 & 89.25 & \underline{1.03}              % PRM*
                         & 309.92 & 88.77 & 88.51 & 0.61 \\     % RRT*
                         
                && \textbf{S-Path(s)}  & 21.93 & \cellcolor{yellow}{89.96} & \cellcolor{yellow}{89.41} & \textbf{4.02}      % BIT*
                         & 27.16 & \cellcolor{yellow}{90.81} & \cellcolor{yellow}{89.60} & \textbf{5.99}       % PRM*
                         & 169.48 & \cellcolor{yellow}{90.04} & \cellcolor{yellow}{89.35} & \textbf{1.11} \\   % RRT*
        \cmidrule{2-15}
        % ====================================
        % RF2P2
        % ====================================
        &\multirow{3}{*}{\textbf{RF2-P2}} & \textbf{I} & 130.57  & 88.40 & 87.52 & 1.00         % BIT*
                         & 226.39  & 90.79 & 88.53 & 1.00        % PRM*
                         & 188.01 & 88.31 & 87.71 & \underline{1.00} \\        % RRT*
                         
                && \textbf{II} & 71.11 & 88.41 & 87.73 & \underline{1.84}          % BIT*
                         & 176.34 & 88.61 & 87.44 & \underline{1.31}           % PRM*
                         & 482.60 & 86.74 & 86.60 & 0.64 \\     % RRT*
                         
                && \textbf{S-Path(s)}  & 13.31 & \cellcolor{yellow}{88.08} & \cellcolor{yellow}{87.86} & \textbf{9.88}      % BIT*
                         & 26.71 & \cellcolor{greenl}{88.72}& \cellcolor{yellow}{87.77} & \textbf{8.65}       % PRM*
                         & 150.65 & \cellcolor{yellow}{88.57} & \cellcolor{yellow}{87.94} & \textbf{2.05} \\   % RRT*
        \cmidrule{2-15}
        % ====================================
        % SF1
        % ====================================
        & \multirow{3}{*}{\textbf{SF1-P1}} & \textbf{I} & 377.03 & 52.39 & 51.79 & 1.00            % BIT*
                         & 974.88 & 52.54 & 52.05 & 1.00          % PRM*
                         & \textgreater 6000.00  & - & - & -  \\    % RRT*
                         
                && \textbf{II}   & 197.55 & 52.46 & 51.56 & \underline{1.90}             % BIT*
                         & 497.24 & 52.54 & 51.70 & \underline{1.95}              % PRM*
                         & \textgreater 6000.00  & - & - & -  \\       % RRT*
                         
                && \textbf{S-Path(s)} & 120.99  & \cellcolor{yellow}{52.30} & \cellcolor{yellow}{51.72} & \textbf{3.12}    % BIT*
                         & 355.26  & \cellcolor{yellow}{52.27} & \cellcolor{yellow}{51.70} & \textbf{2.74}    % PRM*
                         & 480.08 & 51.86 & 51.38 & - \\           % RRT*
        \cmidrule{2-15}
        % ====================================
        % SF2 P1
        % ====================================
        &\multirow{3}{*}{\textbf{SF2-P1}} & $\textbf{I}$ & 380.36 & 30.06 & 29.01 & 1.00        % BIT*
                 & 3036.94 & 31.71 & 30.97 & 1.00     % PRM*
                 & \textgreater6000.00 & - & - & - \\                                        % RRT*
                 
        && $\textbf{II}$ & 122.57  & 31.07 & 30.24 & \underline{3.13}       % BIT*
                 & 220.35  & 31.58 & 30.46 & \underline{13.61}        % PRM*       
                 & \textgreater6000.00 & - & - & - \\       % RRT*
                 
        && \textbf{S-Path(s)}  & 98.27 & \cellcolor{yellow}{29.95} & \cellcolor{yellow}{29.65} & \textbf{3.97}  % BIT*
                 & 111.70 & \cellcolor{red}{32.95} & \cellcolor{greenl}{30.06} & \textbf{25.40}     % PRM*   
                 & 284.09 & 29.88 & 28.87 & - \\                % RRT*
        \cmidrule{2-15}
        % ====================================
        % SF2 P2
        % ====================================
        &\multirow{3}{*}{\textbf{SF2-P2}} & $\textbf{I}$ & 830.98 & 68.27 & 65.31 & 1.00          % BIT*
                & 3076.78 & 68.86 & 67.15 & 1.00          % PRM*
                & \textgreater 6000.00  & - & - & -  \\                                                     % RRT*
                 
        && $\textbf{II}$ & 494.74 & 66.38 & 64.67 & \underline{1.71}          % BIT*
                 & 680.71 & 66.85 & 64.96 & \underline{4.09}          % PRM*
                 & \textgreater 6000.00  & - & - & -  \\                                                     % RRT*
                 
        && \textbf{S-Path(s)}  & 284.09 & \cellcolor{greenl}{65.79} & \cellcolor{yellow}{65.26} & \textbf{3.03}  % BIT*
                 & 309.19 & \cellcolor{greenl}{66.34} & \cellcolor{greenl}{65.31} & \textbf{4.56}    % PRM*
                 & 1310.23  & 65.38 & 64.95 & - \\              % RRT*
        % \cmidrule{2-15}
        % ====================================
        % SF2 P3
        % ====================================
        % &\multirow{3}{*}{\textbf{SF2-P3}} & $\textbf{I}$ & 1146.81 & 68.61 & 66.89 & 1.00          % BIT*
        %         & 4196.83 & 70.47 & 68.57 & 1.00          % PRM*
        %         & \textgreater6000.00  & - & - & -  \\                                                     % RRT*
                 
        % && $\textbf{II}$ & 699.62 & 66.64 & 65.77 & \underline{1.66}          % BIT*
        %          & 1525.88 & 67.09 & 66.09 & \underline{2.79}          % PRM*
        %          & \textgreater6000.00  & - & - & -  \\                                                     % RRT*
                 
        % && \textbf{S-Path(s)}  & 404.33 & \cellcolor{greenl}{66.15} & \cellcolor{greenl}{65.66} & \textbf{2.89}  % BIT*
        %          & 1163.57 & \cellcolor{greenl}{66.67} & \cellcolor{greenl}{66.18} & \textbf{3.68}    % PRM*
        %          & 2045.37  & 65.56 & 65.34 & - \\              % RRT*
        \midrule\midrule
        % ====================================
        % SF2 - Re-planning Scenario #1
        % ====================================
        \multirow{14}{*}{\rotatebox{90}{\textit{Re-planning Scenarios}}} & \multirow{4}{*}{\textbf{SF2-RS1}} 
        & $\textbf{I}$ & 3631.95  & 69.53 & 66.85 & 1.00         % BIT*
                 & 9624.65  & 78.88 & 74.41 & 1.00        % PRM*
                 & \textgreater 10000.00 & - & - & - \\        % RRT*
                 
        && $\textbf{II}$ & 800.33 & 66.84 & 66.14 & 4.67          % BIT*
                 & 2830.32 & 66.69 & 66.26 & 3.58           % PRM*
                 & \textgreater 10000.00 & - & - & - \\     % RRT*
                 
        && $\textbf{III}$  & 467.46 & 66.53 & 66.21 & \underline{8.04}      % BIT*
                 & 1088.09 & 67.04 & 66.23 & \underline{9.26}       % PRM*
                 & 2830.32 & 66.57 & 66.57 & - \\   % RRT*
        
        && \textbf{S-Path(s)}  & 458.61 & \cellcolor{greenl}{66.49} & \cellcolor{yellow}{66.27} & \textbf{8.21}      % BIT*
                 & 672.77 & \cellcolor{greenl}{67.18} & \cellcolor{greenl}{66.21} & \textbf{14.95}       % PRM*
                 & 2731.59 & 66.06 & 65.86 & - \\   % RRT*
        \cmidrule{2-15}
        % ====================================
        % SF2 - Re-planning Scenario #2
        % ====================================
        &\multirow{4}{*}{\textbf{SF2-RS2}} 
        & $\textbf{I}$ & 1329.12  & 33.54 & 33.31 & 1.00         % BIT*
                 & 3897.78  & 34.41 & 33.85 & 1.00        % PRM*
                 & \textgreater 10000.00 & - & - & - \\        % RRT*
                 
        && $\textbf{II}$ & 457.66 & 33.64 & 33.25 & 2.89          % BIT*
                 & 736.55 & 34.39 & 33.27 & 5.20           % PRM*
                 & 9699.72 & 33.48 & 33.36 & - \\     % RRT*
                 
        && $\textbf{III}$  & 367.32 & 33.61 & 33.39 & \underline{3.64}      % BIT*
                 & 410.11 & 34.06 & 33.32 & \underline{9.45}       % PRM*
                 & 864.12 & 33.26 & 33.11 & - \\   % RRT*
        
        && \textbf{S-Path(s)} & 25.00 & \cellcolor{red}{34.51} & \cellcolor{yellow}{34.03} & \textbf{52.79}      % BIT*
                 & 318.42 & \cellcolor{yellow}{34.28} & \cellcolor{yellow}{34.01} & \textbf{12.35}       % PRM*
                 & 462.41 & 34.06 & 33.92 & - \\   % RRT*
        \cmidrule{2-15}
        % ====================================
        % SF2 - Re-planning Scenario #3
        % ====================================
        &\multirow{4}{*}{\textbf{SF2-RS3}} 
        & \textbf{I} & 1820.57  & 67.38 & 46.22 & 1.00 
                    & 9061.63  & 79.60 & 55.38 & 1.00 
                    & \textgreater10000.00 & - & - & - \\
                 
        && \textbf{II} & 800.33 & 46.78 & 46.00 & 3.26
                 & 2896.14 & 46.75 & 46.23 & 4.47           % PRM*
                 & \textgreater 10000.00 & - & - & - \\     % RRT*
                 
        && \textbf{III}  & 458.61 & 46.46 & 46.08 & \underline{5.74}      % BIT*
                 & 800.33 & 47.15 & 46.10 & \underline{15.91}       % PRM*
                 & 4929.15 & 45.76 & 45.66 & - \\   % RRT*
        
        && \textbf{S-Path(s)} & 455.50 & \cellcolor{greenl}{46.40} & \cellcolor{yellow}{46.20} & \textbf{5.80}      % BIT*
                 & 566.47 & \cellcolor{greenl}{46.71} & \cellcolor{greenl}{46.22} & \textbf{22.75}       % PRM*
                 & 2160.89 & 46.35 & 46.22 & - \\   % RRT*
        \bottomrule
    \end{tabular}
    }
    \label{tab:summary_efficiency}
    \vspace{-13pt}
\end{table*}

\smallskip
\noindent \textbf{Parallel execution methodology }
This aspect is evaluated by enabling multithreading in the sequential S-Path(s), resulting in the proposed S-Path.
% to further demonstrate the reduction in $ttp$ achieved by S-Path.
A total of 77 random queries are evaluated, distributed across all environments, with each query consisting of between 2 and 10 subproblems. 
The evaluation metric is the \textit{speedup} factor, defined as the ratio of the time required to solve the subproblems sequentially (using S-Path(s)) versus in parallel (using S-Path). 
For these experiments, \(ttp\) is set to a constant value of \(1000\,\mathrm{ms}\) \textcolor{ black}{(including the time needed to distribute subproblems to parallel workers)}, and PRM* is used as the underlying sampling-based planner.

\smallskip
\noindent \textbf{Hardware} All experiments were conducted on a laptop equipped with a 12th Gen Intel(R) Core(TM) i9-12900H processor that has 14 cores with $m=\;$20 threads.

\begin{figure}   
    \centering
    \includegraphics[width=0.9\linewidth]{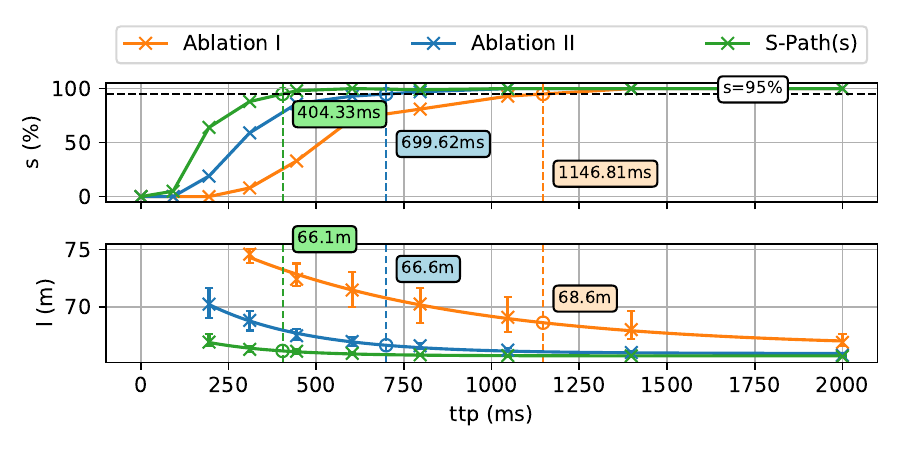}
    \caption{Success rate $s$ and path length $l$ with respect to $ttp$ values across different methods on the SF1-P1 scenario with BIT*. Annotated values indicate $ttp_{95\%}$ and $l_{95\%}$, respectively. }
    \label{fig:success_rate}
    \vspace{-5pt}
\end{figure}

\vspace{-4pt}
\subsection{Results and Discussions}
\noindent \textbf{Sequential execution results }
\color{ black}
\Cref{tab:summary_efficiency} summarizes the efficiency gains of the evaluated methods across all queries. 
Specifically, in the planning scenarios, S-Path(s) demonstrates multifold gains in efficiency, with an average increase of \(6.0\times\).
This is even without considering its pairing with RRT*, where in many cases S-Path(s) converges while other ablations do not. 
The efficiency gains from pairing with PRM* are greater (\(8.7\times\) on average) than those from pairing with BIT* (\(4.5\times\) on average), because the uninformed PRM* benefits more from the C-space restrictions provided by S-Path(s).
Moreover, while S-Path(s) on average produces slightly longer paths (difference in $l_{95\%}$ between $-5.0\%$ and $+15.9\%$, with an average of $+2.5\%$), due to its constraint of passing through the center of doorways; 
this is compensated for by substantial improvements in the \textit{time-to-plan} (reduction in $ttp_{95\%}$ between $1.1\times$ and $27.2\times$, with an average of $6.0\times$). 
\color{black}

In replanning scenarios, S-Path consistently produces shorter paths, averaging $14.9\%$ shorter than those generated by \textbf{I}.
Its heuristics prove especially beneficial in environments with blockages, where increased complexity frequently causes traditional heuristic-guided planners to fail. 
Furthermore, the ability to cache and reuse previously solved subproblems eliminates the need for redundant computations, further reducing $ttp$ and improving overall efficiency, by up to $52\times$ for BIT* and $22\times$ for PRM*.
% some space can be gained below
A notable example is the SF2-RS2 with BIT*, where the task becomes nearly trivial, as most subproblems are either cached from the initial planning attempt or are reduced to simple point-to-point queries without obstacles.
% , that are trivial for BIT*.

Furthermore, each individual component of S-Path contributes incrementally to the overall efficiency gains (see \Cref{fig:success_rate} for an example run).
The C-space restriction introduced in \textbf{II} provides the most significant improvement, as it limits sampling to only relevant regions, allowing planners to make more progress within the same computational time.
Further efficiency gains are achieved through subproblem decomposition in S-Path (or \textbf{III} in the \textit{Replanning Scenarios}), although these gains are still limited by the sequential execution setup (the results with parallel execution that follow demonstrate additional speedups).
An exception was observed in the RRT* experiments on the RF2 environment, where the C-space restriction introduced in \textbf{II} leads to performance regressions, likely due to interference with RRT*'s random sampling strategy. 
Despite this, S-Path(s) manages to converge by decomposing the global planning problem into smaller subproblems, thereby mitigating the negative impact of the C-space restriction.
Finally, the reuse of previously solved subproblems in the \textit{Replanning Scenarios} leads to further improvements over \textbf{III}, particularly for RRT*, which otherwise incurs high computational costs due to its random sampling.

\begin{figure}   
    \centering
    \includegraphics[width=0.80\linewidth]{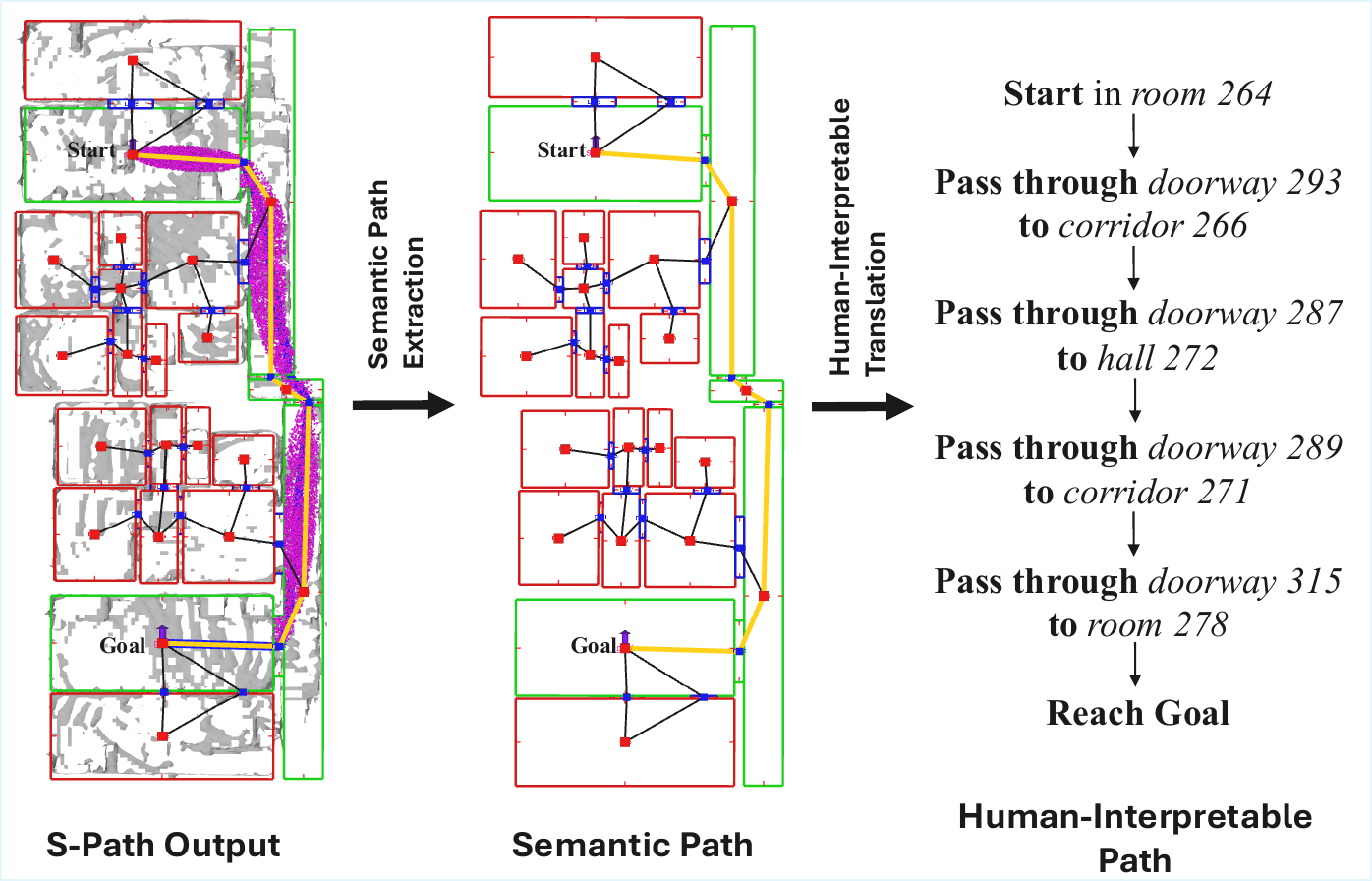}
    \caption{Example of human-interpretable path generation from S-Path output. First, the semantic path is extracted, which is then translated into human-interpretable commands.}
    \label{fig:human_interpretable}
    \vspace{-5pt}
\end{figure}

Overall, S-Path substantially benefits underlying planners that lack built-in heuristics, such as RRT*, while also enhancing the performance of informed planners like BIT*. 
Additionally, S-Path produces interpretable plans by utilizing semantic concepts such as rooms and doorways (see \Cref{fig:human_interpretable}).

\begin{figure}   
    \centering
    \includegraphics[width=0.75\linewidth]{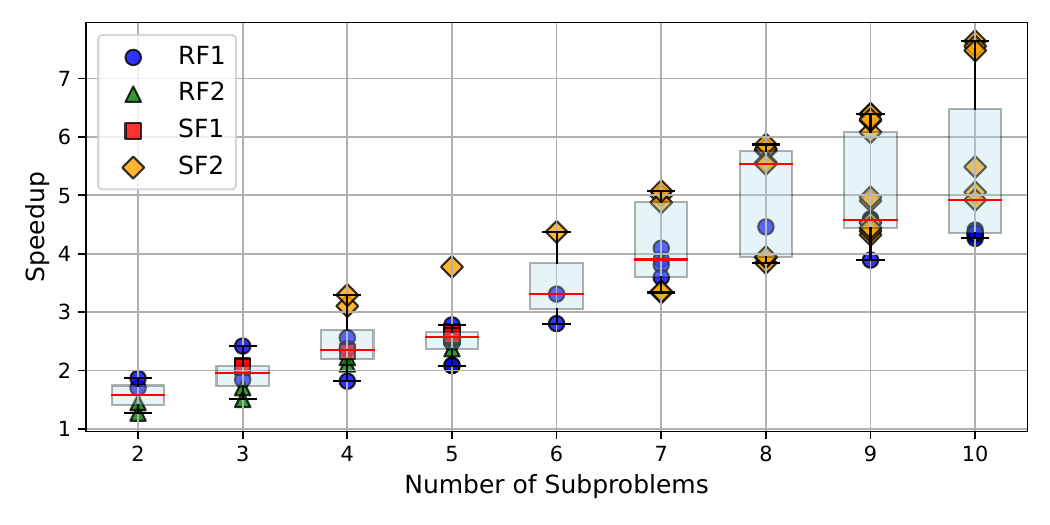}
    \caption{Parallel execution speedups versus the number of identified subproblems, evaluated across different environments. Each marker corresponds to a single query. Overall, the speedup increases approximately proportionally with the number of subproblems, with synthetic environments showing higher speedups due to the absence of irregular obstacles.}
    \label{fig:speedup}
    \vspace{-5pt}
\end{figure}

\smallskip
\noindent \textbf{Parallel execution results }
S-Path yields parallel execution speedups that correlate with the number of subproblems, as shown in \Cref{fig:speedup}, though the actual gains depend heavily on the subproblem setup, since the subproblems vary in complexity. 
When sufficient cores are available to process all subproblems concurrently, the total parallel execution time approaches the \(ttp\) of the largest subproblem. 
This results in greater variation in speedups for queries with more subproblems, as emphasized by the taller boxplots in \Cref{fig:speedup}. 
Conversely, when there are fewer cores than subproblems, the speedup gains diminish accordingly.
However, in all queries, including those in large environments such as SF2, the number of subproblems was at most 10, so the number of available cores was not a limiting factor.
This is largely due to subproblem merging, which avoids the creation of very small, practically inefficient subproblems (as discussed in \Cref{sec:subproblem-solver}), making S-Path highly scalable. 
Furthermore, it is important to note that the semantic planning for all queries took less than \(1\,\mathrm{ms}\), making its overhead negligible.

\vspace{-3pt}
\section{Conclusion} 
\label{sec:conclusion}
In this work, we proposed S-Path, a novel metric-semantic path planning algorithm that augments sampling-based planners with semantic information derived from indoor 3D scene graphs, which improves planning efficiency while providing human-interpretable path planning. 
Experimental results demonstrate that \spath achieves up to $25\times$ efficiency improvement over the underlying sampling-based planner. 
This originates from the decomposition of the global planning problem into multiple independent subproblems by exploiting room and doorway semantics, effectively reducing the configuration space.
This decomposition also enables parallel execution, allowing for further speedup which correlate with the number of subproblems.
\spath can also efficiently replan by reusing previously solved subproblems, achieving up to $52\times$ overall efficiency improvement. 
% In scenarios where the original path becomes infeasible due to blockages, \spath can efficiently replan by reusing previously solved subproblems, achieving up to $52\times$ overall efficiency improvement. 
However, \spath enforces traversal through the center of doorways, which may lead to suboptimal paths when considering wide doorways. 
Additionally, \spath currently assumes rooms delimited by planar walls, which might arise challenges in contour generation and sample space restriction in environments with complex curved architectural geometries. 
% To address these limitations, 
\textcolor{ black}{Moreover, S-Path assumes an accurate map of the environment initially provided and does not account for uncertainties.}
Future work will focus on refining path optimality 
through a local-to-global path optimization step, 
extending contour generation to handle more complex architectural structures, \textcolor{ black}{and incorporating robustness to uncertainties in the environment.}

\vspace{-3pt}
\bibliographystyle{IEEEtran}
{\footnotesize
\bibliography{main}}

\end{document}